\documentclass{article}

   \usepackage[final]{neurips_2020}

\usepackage[utf8]{inputenc} 
\usepackage[T1]{fontenc}  
\usepackage{hyperref}    
\usepackage{url}      
\usepackage{booktabs}    
\usepackage{amsfonts}    
\usepackage{microtype}   
\usepackage{graphicx}
\usepackage{caption,subcaption}
\usepackage{amsmath}
\usepackage{tabularx,booktabs}
\usepackage[flushleft]{threeparttable}
\usepackage{algorithm}
\usepackage[noend]{algpseudocode}
\usepackage{float}

\title{Self-Supervised Learning for Fine-Grained Image Classification}

\author{  Farha Al Breiki \\ (20020015@mbzuai.ac.ae) \And Muhammad Ridzuan \\ (20020084@mbzuai.ac.ae) \And Rushali Grandhe \\ (20020076@mbzuai.ac.ae)
}

\begin{document}

\maketitle

\section{Abstract}

Fine-grained image classification involves identifying different subcategories of a class which possess very subtle discriminatory features. Fine-grained datasets usually provide bounding box annotations along with class labels to aid the process of classification. However, building large scale datasets with such annotations is a mammoth task. Moreover, this extensive annotation is time-consuming and often requires expertise, which is a huge bottleneck in building large datasets. On the other hand, self-supervised learning (SSL) exploits the freely available data to generate supervisory signals which act as labels. The features learnt by performing some pretext tasks on huge unlabelled data proves to be very helpful for multiple downstream tasks.

Our idea is to leverage self-supervision such that the model learns useful representations of fine-grained image classes. We experimented with 3 kinds of models: Jigsaw solving as pretext task, adversarial learning (SRGAN) and contrastive learning based (SimCLR) model. The learned features are used for downstream tasks such as fine-grained image classification. Our code is available at \url{https://github.com/rush2406/Self-Supervised-Learning-for-Fine-grained-Image-Classification}.

\section{Dataset}

We used the fine-grained cassava plant disease dataset available at \url{https://www.kaggle.com/c/cassava-disease/data}. It consists of 12k unlabelled images and 6k labelled images in .jpg format, divided among five subcategories according to plant disease type: cbb (cassava bacterial blight), cbsd (cassava brown streak disease), cgm (cassava green mite), cmd (cassava mosaic disease) and healthy. The images are varied in size, lighting, background, and resolution.

The unlabelled images were used during the SSL pretext task and labelled images were used during the downstream task. The labelled images had an imbalance in the number of images per class. This was handled by obtaining more images from another publicly available dataset and by using class weights. We randomly split the dataset into 80\% training and 20\% validation.

\section{Related Work}
Fine-grained image classification is a very well-known problem in the computer vision domain. Deep neural networks have been used extensively for fine-grained classification. Some of the most popular ones include kernel pooling with CNNs ~\cite{8099808} and boosted CNNs ~\cite{BMVC2016_24}. Over the years, several approaches have been developed - from requiring direct human supervision to weakly supervised approaches. Methods like ~\cite{7298658,493,huang2015partstacked,6618954,zhang2014partbased} require a lot of part-based annotation for the datasets.

Recently, weakly supervised methods requiring only image level labels have gained a lot of popularity. There have been numerous attention-based approaches like ~\cite{xiao2014application,8237819}. Moreover, DCL ~\cite{8953746}, and PMG ~\cite{du2020finegrained} methods have shown remarkable performance without explicit attention.

Self-supervised learning has received a huge surge in recent years owing to the difficulties posed by supervised learning methods. Expensive annotations, generalization error, spurious correlations, and adversarial attacks are examples of the issues associated with supervised learning ~\cite{jing2019selfsupervised}. Self-supervised learning involves learning features by performing a pretext task. Pretext tasks like generation-based, context-based, free semantic label-based and cross modal-based have been used to learn discriminatory features ~\cite{jing2019selfsupervised, noroozi2017unsupervised, larsson2017colorization}. These learned features are utilized to complete the downstream task, which can be classification, detection, segmentation and others.

According to ~\cite{Liu_2021}, self-supervised learning can be classified into three main categories: generative learning, contrastive learning, or generative-contrastive (adversarial) learning. The main difference between them is their architecture and objective. 

Generative self-supervised models can be autoregressive (AR) models, flow-based models, auto-encoding (AE) models, and hybrid generative models ~\cite{Liu_2021}. Generative models have been popular in the NLP domain for text classification. BERT, MADE, and CBOW are some of the well-known applications for generative-based models ~\cite{devlin2019bert,germain2015made,mikolov2013efficient}. 

Contrastive learning has been extensively employed for self-supervision ~\cite{chen2020simple,he2020momentum}. Unlike generative models, contrastive models try to reduce the dissimilarity between augmentations of the an image~\cite{Liu_2021}. Models such as Deep InfoMax, MoCo, and SimCLR have been used for self-supervised classification applications in ~\cite{hjelm2019learning,he2020momentum, chen2020simple} . 

Adversarial learning combines some generative and contrastive learning features as it learns to reconstruct the original data distribution by minimizing the distributional divergence. Original data can be constructed by feeding complete input or partial input. Applications for adversarial learning with complete input are BiGAN and ALI ~\cite{donahue2017adversarial, dumoulin2017adversarially}, and for partial input, we have colorization, inpainting, and super-resolution. 

Self supervised learning for fine-grain hasn't been explored much yet. In this project, we have tried a pretext task,adversarial learning and contrastive learning for self-supervised fine-grain classification. We have used Jigsaw solving as pretext task. In adversarial learning, we used a generation-based (super-resolution) task that was based on SRGAN ~\cite{ledig2017photorealistic}. Further, for contrastive learning, we have experimented using the SimCLR model with different augmentations such as patch swapping, coarse dropout and jigsaw.

\section{Method}

\subsection{Baseline model}

We chose a weakly supervised fine-grained classification model -  Fine-Grained Visual Classification via Progressive Multi-Granularity Training of Jigsaw Patches ~\cite{du2020finegrained}. It progresses from a high granularity to a low granularity, thus combining low-level specific patch details along with the high-level view of the entire input image.  It uses image patches of increasing sizes to learn the characteristic features in a step-by-step manner. The ResNet model achieved an accuracy of 88\% on the labelled images.

\subsection{Jigsaw as Pretext Task}

The  idea  of  using  Jigsaw  solving  as  a  pretext  task ~\cite{noroozi2017unsupervised}  comes  from  the  observation  that  solving  a Jigsaw puzzle not only requires observing the individual patches but also understanding the spatial relationships between the patches. This in turn demands learning specific discriminatory features of the patches that can help in solving the puzzle. Fine-grained classes usually have very subtle distinguishing features. Hence, we explored on how well the model could learn these fine features through solving the Jigsaw puzzle.\\

The input image is split into 3x3, i.e. 9 patches, which are used to create a jigsaw. Permutations for shuffling the patches are generated such that the average Hamming distance between the permutations is the maximum. Moreover, multiple jigsaws are generated for each input image. These are done to ensure that the model does not learn any sort of shortcuts to predict the right order and also makes every position equally likely for every patch. The model is tasked with predicting the right permutation that solves the jigsaw. The learnt features were used for downstream fine-grained classification. We have fine-tuned using different number of layers as results are summarized in Table \ref{tab:jig_pretext}. Downstream accuracy of 67\% was obtained. 

\subsection{Contrastive Learning: self-supervised learning using SimCLR} \label{sec:aug}

SimCLR is a self-supervised learning algorithm based on contrastive loss. It generates two augmented images of each image and ensures that the learned representations for those two images are closer to each other in the feature space but are farther away from the representations of the other images of the batch. To achieve this purpose, SimCLR uses the NT-Xent, or Normalized Temperature-scaled Cross Entropy Loss as a loss function. Let $z_i$ and $z_j$ be the feature representations of the same altered image. The loss function for a positive pair of examples (i, j) is defined as:

\begin{equation}\label{eq:nt_xent}
\\  l_{i,j} = -log\frac{exp(sim(z_i,z_j)/\tau)}{\displaystyle\sum_{k=1}^{2N}\mathbf{1}_{[k\neq i]}exp(sim(z_i,z_k)/\tau)}
\end{equation}

The default augmentations used in SimCLR are random resize cropping, color jittering, random horizontal flip, random grayscale and Gaussian blurring. 
Considering the success of SimCLR, we decided to experiment and observe the performance of the model on fine-grained datasets.

We used the default augmentations of SimCLR (mentioned above), except Gaussian blurring (as it would also blur the discriminatory spots which are essential to be learnt). After the contrastive learning phase, we performed the downstream classification task and achieved an accuracy of 73\%. The results have been summarized in Table \ref{tab:original} with a sample Grad-CAM shown in Figure \ref{fig:c_origin}. It can be inferred from the Grad-CAM that the model had used the spots on the leaf to make the class prediction i.e. the model was able to localize the fine-grained features to a decent extent. This motivated us to experiment with some modifications to the SimCLR pipeline so that the fine-grained discriminatory features may be learnt better.

\subsubsection{Gamma transform}

The first image-level augmentation that we used is a random gamma transform from the albumentations library ~\cite{Buslaev_2020}. This transform changes the brightness of the image as shown in Figure \ref{fig:gam}. The brightness is altered randomly within a range of 50-250. Level 50 refers to a darker image and 250 corresponds to a very bright image. The gamma transform was performed before the default SimCLR augmentations.

\textbf{NOTE:} In the default augmentations, random resize cropping was used. We realized that this may not be suitable because the random crop may or may not contain the essential fine-grained regions. Hence, we replaced it with resize for the following augmentations.

\subsubsection{Coarse Dropout}
Coarse dropout was the second image-level augmentation that was applied from the albumentations library ~\cite{Buslaev_2020}. This transform randomly removes squares from the input image. The size of the squares ranges from 10 to 25 pixels. Figure \ref{fig:drop} shows an example of the output of the augmentation. Coarse dropout is generally used to prevent overfitting while using images. Hence, this motivated us to check if the model was probably overfitting and verify if it could still learn the features despite the removed regions. We tried to make sure that the fine-grained regions are not completely lost while using this transform by controlling the size and number of squares.

\subsubsection{Random Patch swapping}

In order to predict the correct class label, it is important that the model is able to localize the fine-grained features. Thus, the model should be able to identify those discriminatory features, irrespective of the position of the features. This intuition prompted us to use a simple random patch swapping component. Two random patches of size 200x200 are extracted and swapped in every image. Thus, training the model leads to the original image Figure \ref{fig:input} and the image with random patches swapped Figure \ref{fig:swap} to have representations closer in the embedding space.

\subsubsection{Random Jigsaw Shuffling} \label{sec:Jigsaw}

Our baseline model ~\cite{du2020finegrained}, Progressive Multi Granularity (PMG) model for fine-grained classification, uses multi granularity Jigsaw shuffling to help improve feature learning. Drawing inspiration from this idea, we incorporated random Jigsaw shuffling into the SimCLR pipeline along with the existing augmentations. A random permutation of [0, 1, 2.., $n \times n$ - 1] is generated, where $n$ is the granularity of the Jigsaw. The image is divided into uniform $n \times n$ partitions and shuffled according to the generated permutation.\\

We performed two such variations. The first comprised of the original image and a randomly-shuffled Jigsaw puzzle (4x4) image. This achieved a downstream task accuracy of 69.5\%. The second variation involved two randomly-shuffled Jigsaw puzzles with (4x4) and (2x2) granularities. This achieved a downstream task accuracy of 68.7\%. The representations of these pairs of images (Figure \ref{fig:Jigsaw}) are learnt to be closer in the embedding space. 

\subsubsection{DCL-based Jigsaw shuffling}

From the results as shown in Table \ref{tab:simCLR}, it can be observed that random Jigsaw shuffling did not perform quite well. This can be attributed to the fact that during the Jigsaw formation (dependent on the granularity), some of the fine-grained regions might be chopped and shuffled such that they are no longer identifiable. To handle this scenario, another algorithm from the fine-grained domain provided a plausible solution - DCL~\cite{8953746}. It suggests that shuffling within the local neighborhood of the image can help preserve the fine-grained regions to a large extent. The algorithm has been summarized in Algorithm \ref{algo:dcl}.

\begin{algorithm}
\caption{\textbf{DCL Jigsaw shuffling}}
\label{algo:dcl}
\begin{algorithmic}[1]

\Procedure{DclJigsaw}{$img,n=7$}\\  

  \State Uniformly partition img into $n \times n$ regions. Each region is denoted as $R_{i,j}$.\\
  
  \For{every $row_j$ in R}
  
    \State Generate random vector $q_j$ such that $q_{j,i}$ = i + r, where $r \sim U(-k,k), 1 \leq k \leq n$ \\
    
    \State Generate new permutation $\sigma_j^{row}$ of regions in $j^{th}$ row by sorting the array $q_j$, verifying:\\
    \State $\forall i \epsilon \{1,...,n\} ,|\sigma_j^{row}(i) - i| < 2k$
  \EndFor\\
  
  \State Similarly, perform the above operation for each $column_i$ in R such that:
  
  \State $\forall j \epsilon \{1,...,n\} ,|\sigma_i^{col}(j) - j| < 2k$\\
  
  \State $R_{i,j}$ in original region location is mapped to a new coordinate: $\sigma(i,j) = (\sigma_j^{row}(i),\sigma_i^{col}(j))$\\
  
  \State Perform shuffling of img based on the new coordinate mapping.\\
\State \textbf{return} img
\EndProcedure
\end{algorithmic}
\end{algorithm}

\begin{figure}[H]
\centering
\begin{subfigure}[b]{.2\linewidth}
\includegraphics[width=\linewidth]{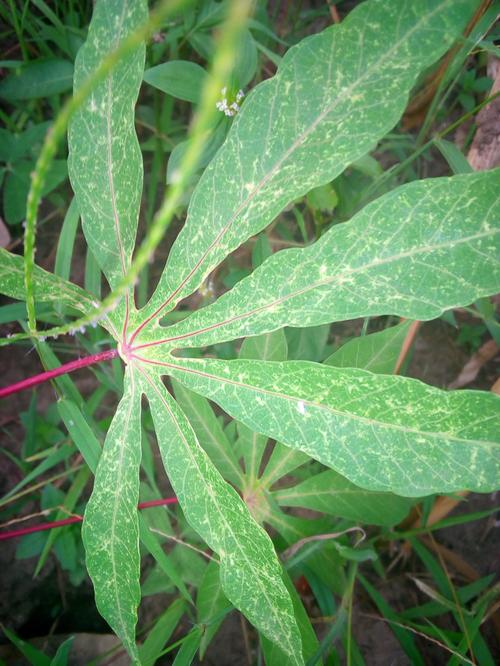}
\caption{Input image}\label{fig:input}
\end{subfigure}
\begin{subfigure}[b]{.2\linewidth}
\includegraphics[width=\linewidth]{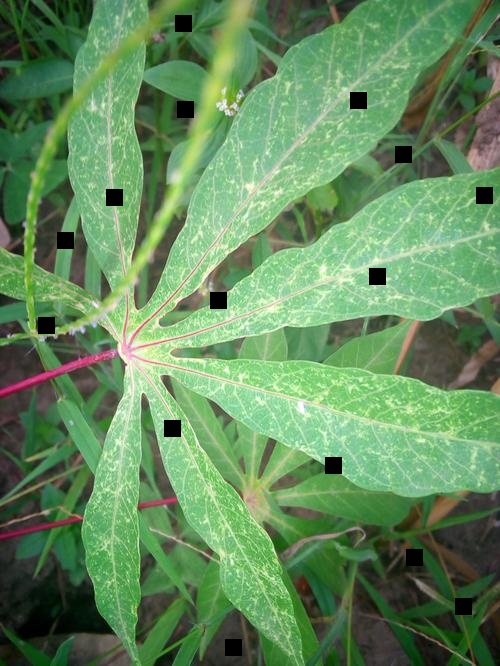}
\caption{Coarse dropout}\label{fig:drop}
\end{subfigure}
\begin{subfigure}[b]{.2\linewidth}
\includegraphics[width=\linewidth]{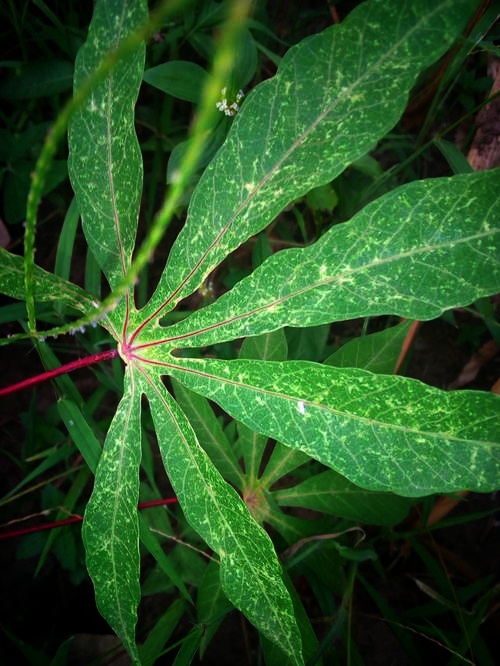}
\caption{Gamma transform}\label{fig:gam}
\end{subfigure}
\begin{subfigure}[b]{.2\linewidth}
\includegraphics[width=\linewidth]{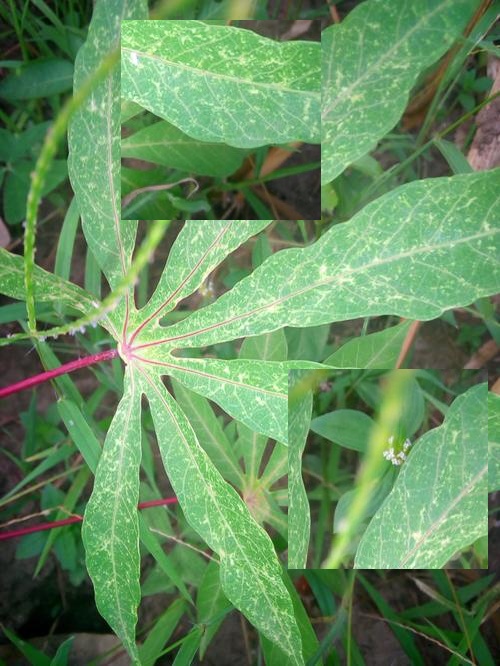}
\caption{Patch swap}\label{fig:swap}
\end{subfigure}
\begin{subfigure}[b]{.2\linewidth}
\includegraphics[width=\linewidth]{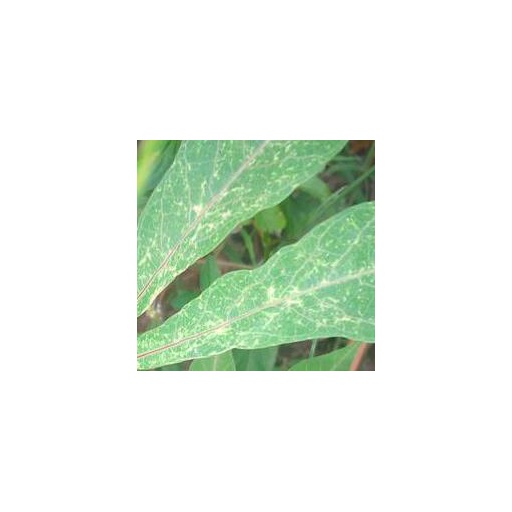}
\caption{fine-grained region}\label{fig:fine}
\end{subfigure}
\begin{subfigure}[b]{.2\linewidth}
\includegraphics[width=\linewidth]{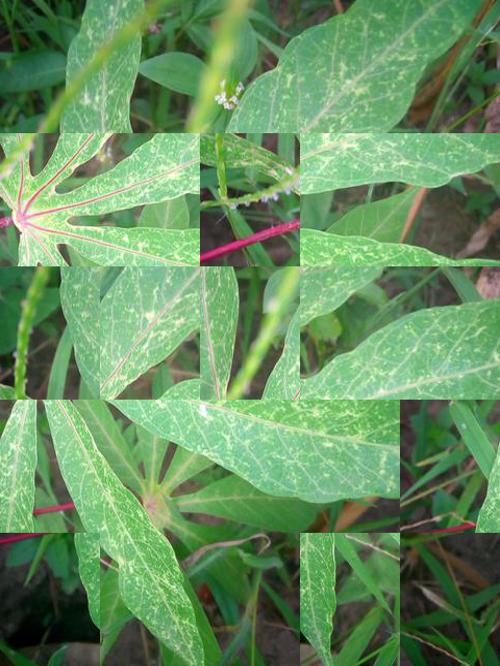}
\caption{Random Jigsaw}\label{fig:Jigsaw}
\end{subfigure}
\begin{subfigure}[b]{.2\linewidth}
\includegraphics[width=\linewidth]{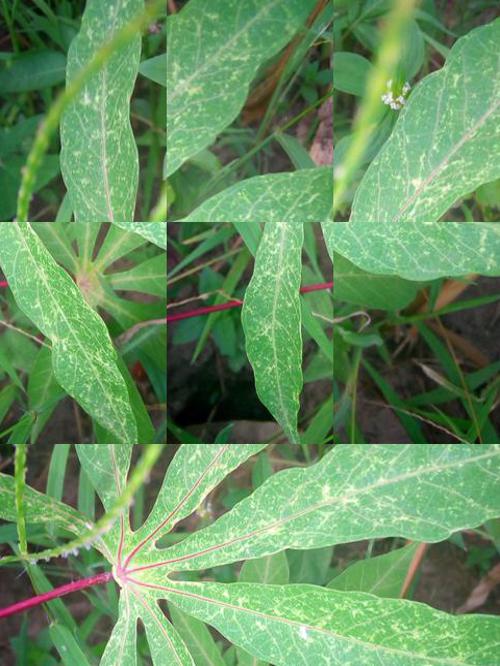}
\caption{DCL Jigsaw}\label{fig:dcl3x3}
\end{subfigure}
\caption{Augmentations experimented on SimCLR as explained in Section \ref{sec:aug}}
\end{figure}

The representations of the original image and DCL-based Jigsaw pair Figure \ref{fig:dcl3x3} are learnt to be closer in the embedding space. We experimented with different granularities of Jigsaw (i.e. 3x3, 5x5, 7x7), out of which 3x3 Jigsaw showed the best accuracy.

\subsubsection{Fine-grained region cropping}

Using bounding box annotations for fine-grained regions have been very popular in the fine-grained datasets. Using this as the intuition, we tried to form image pairs involving the original image and a crop of the most important fine-grained region in the image which are supposed to be resulting in representations closer in the embedding space. We carried out this experiment on a small scale by generating bounding boxes for 200 images (40 images for each class). The results looked promising and hence, we tried to perform the same on the larger dataset.

As the dataset did not have any bounding box annotations, we employed a technique of smartcropping to obtain the fine-grained regions. Smartcrop is a way of intelligently determining the most important part of the image and keeping it in focus while cropping the image. The tool was able to approximately localize the fine-grained regions to a good extent. The smartcrop algorithm can be summarized by the following steps - \\

1. Find edges using Laplace\\
2. Find and boost regions high in saturation\\
3. Generate a set of candidate crops using a sliding window\\
4. Rank them using an importance function to focus the detail in the center and avoid it in the edges.\\
5. Output the candidate crop with the highest rank\\

We have directly used a smartcropping package available on Github~\cite{SmartCrop}. After localizing the fine-grained regions, we overlaid them against a white background Figure \ref{fig:fine} to maintain the image size.

\subsection{Scaling: Self-supervised learning using SRGAN}
Scaling was chosen as a pretext task to allow the model to learn the finer features of the images since the dataset images were obtained using phone cameras of varied quality and resolutions. In ~\cite{ledig2017photorealistic}, Ledig et al. developed a super-resolution generative adversarial network (SRGAN) to create high resolution images. The network downscales the images by a factor of four using a bicubic kernel and regenerates the images at two, four, and eight times the resolution.\\

Following the work for the midterm where we investigated the potential of the SRGAN discriminator as a feature extractor, we now proceeded to replicate the same procedure with the SRGAN generator.
The intuition behind this approach is that as the network upsamples an input image, it also learns the finer features of the image that are important to distinguish the fine-grained classes of the dataset.\\

Traditionally, an MSE content loss has been calculated to ensure the preservation of pixel-level content (eq. \ref{eq:mse_content_loss}) ~\cite{dong2015image, shi2016realtime}. However, the resulting upsampled image tends to be overly smoothed and lacks finer details. ~\cite{ledig2017photorealistic} introduced a VGG content loss instead, where the focus of the loss function optimization is shifted from the pixel space to the feature space to ensure high-level content preservation. This is achieved by calculating MSE comparing the generated images ($G_{\theta_G}(I^{LR}$)) with the feature maps obtained after the $j$-th convolution before the $i$-th maxpooling layer of the pre-trained VGG network ($\phi_{i,j}(I^{HR}$)) (eq. \ref{eq:vgg_content_loss}).

\begin{equation} \label{eq:mse_content_loss}
\\l_X^{SR} = l_{MSE}^{SR} = \frac{1}{r^2WH}{\displaystyle\sum_{x=1}^{rW}}{\displaystyle\sum_{y=1}^{rH}}(I_{x,y}^{HR} - G_{\theta_G}(I^{LR})_{x,y})^2
\end{equation}

\begin{equation} \label{eq:vgg_content_loss}
\\l_X^{SR} = l_{VGG_{i,j}}^{SR} = \frac{1}{W_{i,j}H_{i,j}}{\displaystyle\sum_{x=1}^{W_{i,j}}}{\displaystyle\sum_{y=1}^{H_{i,j}}}(\phi_{i,j}(I_{HR})_{x,y} - \phi_{I,j}(G_{\theta_G}(I^{LR})_{x,y})^2
\end{equation}

They eventually introduced a perceptual loss ($l^{SR}$), defined as the weighted sum of the content loss ($l_X^{SR}$) and adversarial loss ($l_{Gen}^{SR}$).(eq. \ref{eq:perceptual_loss}).
\begin{equation} \label{eq:perceptual_loss}
\\l^{SR} = l_X^{SR} + 10^{-3}l_{Gen}^{SR}
\end{equation}

\begin{figure}[H]
\centering
\includegraphics[width=\linewidth]{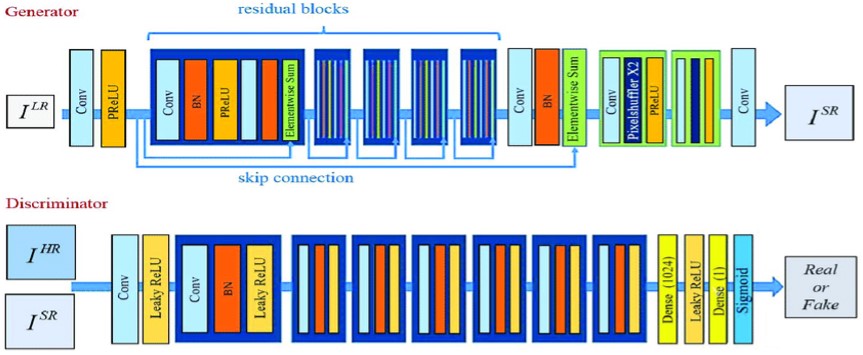}
\caption{SRGAN architecture. The generator takes a low-resolution input through ResNet16 and upsampling block pipeline to output a high resolution image (top). The discriminator tries to distinguish the generated super-resolution image from the original high resolution image by passing them through a VGG backbone architecture (bottom). Reprinted from ~\cite{8917633}.}
\label{fig:srgan_arch}
\end{figure}

The SRGAN generator consists of two main components: a ResNet backbone and an upsampling block. The ResNet consists of 16 identical residual blocks (with convolution-batch normalization-parametric ReLU sequence) and a bypass skip connection that relieves the network from modeling the identity transformation. 
The upsampling block consists of a convolution, pixel shuffler, and parametric ReLU (Figure \ref{fig:srgan_arch}). \\

\begin{figure}[H]
\centering
\includegraphics[width=200pt]{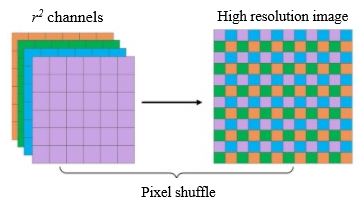}
\caption{Upsampling pixel shuffle operation that converts a tensor of shape ($W, H, C \times r^2$) to a tensor of shape ($W\times r, H\times r, C$), where $r$ is the upscaling factor.}\label{fig:pixel_shuffle}
\end{figure}

The pixel shuffler is the main operation responsible for the upsampling. It acts as a deconvolution operation that reverses the convolution transformation to produce a higher resolution output. This is done through a periodic reshuffling of the lower resolution feature maps into a higher resolution output, essentially going from a tensor of shape ($W, H, C \times r^2$) to a tensor of shape ($W\times r, H\times r, C$), where $r$ is the upscaling factor (Figure \ref{fig:pixel_shuffle}).\\

In our experiments, we retrained the model using the cassava dataset and reconstructed the images at the original resolution (i.e. four times upscaling). To leverage the power of self-supervised learning, we included all 12k unlabelled images in our training. We then isolated the generator from the network, froze the weights, and attached a classifier to output five classes using an L2-Adam regularization and the log softmax activation function. Following the experimental procedure we performed with the discriminator, we changed the input image sizes, reduced the depth of the ResNet backbone, and added dropouts to the fully connected layer of the classifier to evaluate the model performance. The experiments were performed in progressive order such that the depth of the architecture was reduced on the best accuracy input image size, and the dropout was introduced on the best accuracy depth. The results are summarized in Table \ref{tab:scaling_obs}.

\section{Results and Discussion}
We chose accuracy as our primary metric of evaluation as it is generally used for evaluating performance on fine grained data. We have also included alternative metrics including precision, recall, and F1 scores in the Appendix for a more comprehensive evaluation.

For qualitative evaluation, we ran gradient-weighted class activation maps (Grad-CAM) as a visualization tool of our models. Grad-CAM provides a way of debugging the model and visually validating that it is “looking” and “activating” at the correct locations of an image. Grad-CAM works by finding a specified convolutional layer in the network and examining the gradient information flowing into that layer. We used Grad-CAM to visualize the regions the model was using to make a class prediction. Some Grad-CAM images are shown in Figures \ref{fig:c_cam} and \ref{fig:srgan_cam}.

\begin{minipage}{\linewidth}
\centering
\captionof{table}{Experimenting with various parameters} \label{tab:jig_pretext} 
\begin{tabular}{ p{1.5in} p{0.85in} p{0.85in}}\toprule[1.5pt]
\bf Parameter & \bf Value & \bf Classification accuracy\\\midrule
No. of permutations & 100 & 0.658\\
for Jigsaw task & 200 x 200 & 0.676\\
\hline \\
Image size & 256 x 256 & 0.664\\
 & 550 & 0.673\\
\hline \\
 No. of last layers used  & 2 & 0.661\\
 for finetuning & 5 & 0.687\\
\bottomrule[1.25pt]
\end {tabular}\par
\bigskip
\end{minipage}

\subsection{Jigsaw as pretext task}

It can be seen from Table \ref{tab:jig_pretext} that the accuracy improved slightly with an increase in the number of permutations. Using a larger image size was also beneficial. Fine-tuning from more layers also helped increase accuracy. 
While  the  model  performs  well  on  the  pretext  task,  it  is  unable  to  perform  similarly  on  the downstream task and reaches an accuracy of around 68\%. This could be because the model could not learn the distinguishing features from the patches which looked very similar.

\subsection{SimCLR}

\begin{figure}[H]
\centering
\begin{subfigure}[b]{.2\linewidth}
\includegraphics[width=\linewidth]{train-cgm-651.jpg}
\caption{Input image}\label{fig:c_input}
\end{subfigure}
\begin{subfigure}[b]{.2\linewidth}
\includegraphics[width=\linewidth]{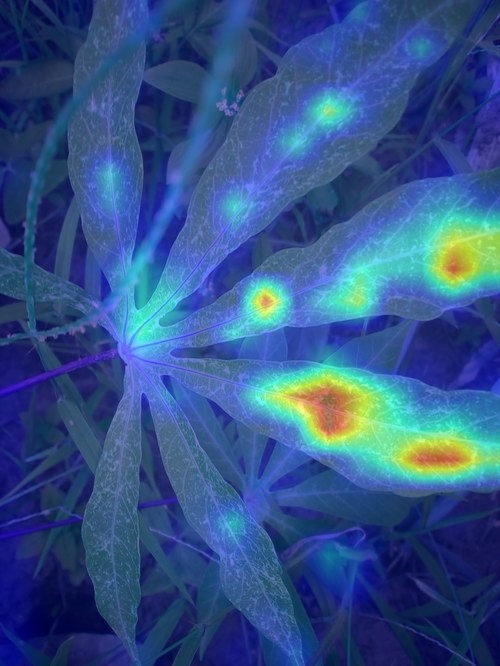}
\caption{SimCLR original}\label{fig:c_origin}
\end{subfigure}
\begin{subfigure}[b]{.2\linewidth}
\includegraphics[width=\linewidth]{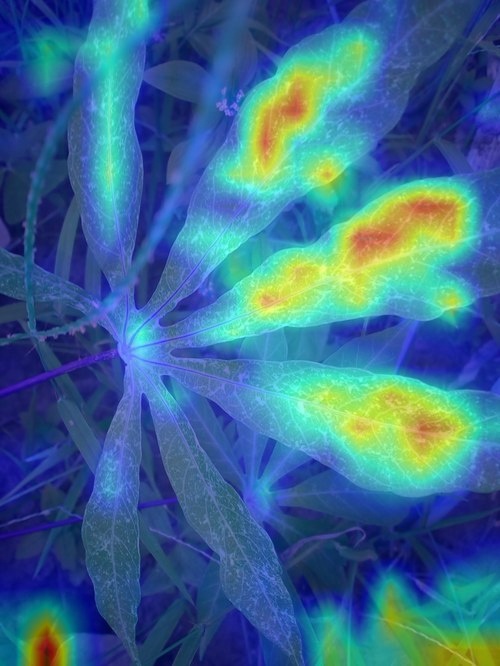}
\caption{Patch swapping}\label{fig:c_swap}
\end{subfigure}
\begin{subfigure}[b]{.2\linewidth}
\includegraphics[width=\linewidth]{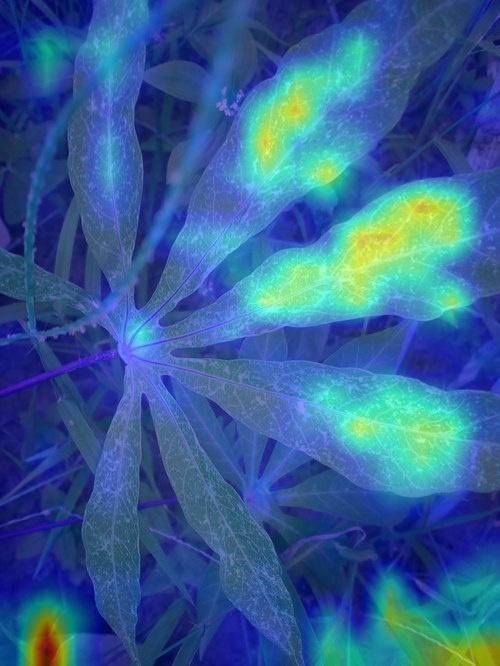}
\caption{Random Jigsaw}\label{fig:c_jigaw}
\end{subfigure}
\begin{subfigure}[b]{.2\linewidth}
\includegraphics[width=\linewidth]{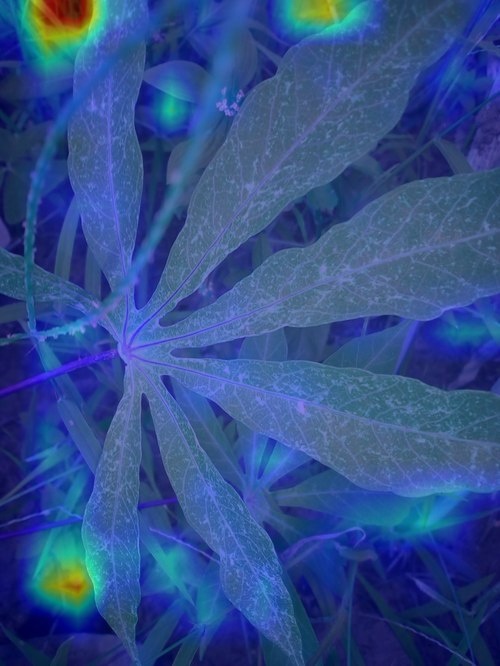}
\caption{Fine-grained crop}\label{fig:c_fine}
\end{subfigure}
\begin{subfigure}[b]{.2\linewidth}
\includegraphics[width=\linewidth]{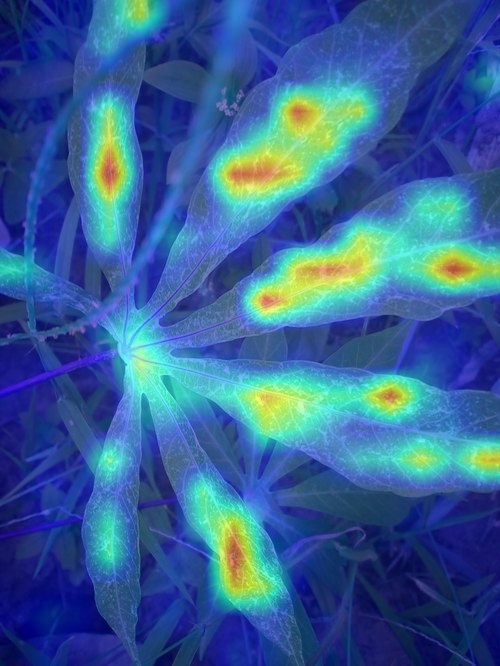}
\caption{DCL 3x3 Jigsaw}\label{fig:c_dcl}
\end{subfigure}
\begin{subfigure}[b]{.2\linewidth}
\includegraphics[width=\linewidth]{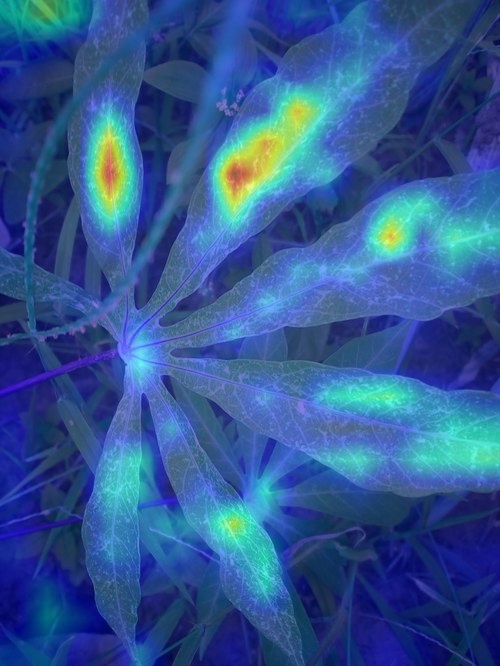}
\caption{Gamma transform}\label{fig:c_gamma}
\end{subfigure}
\begin{subfigure}[b]{.2\linewidth}
\includegraphics[width=\linewidth]{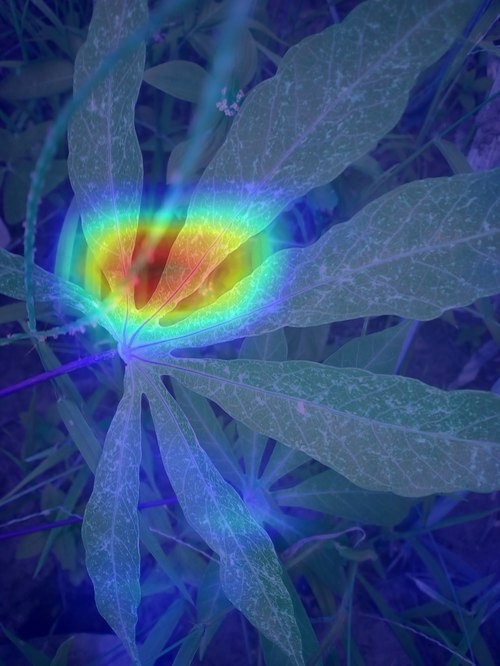}
\caption{Coarse dropout}\label{fig:c_dropout}
\end{subfigure}
\caption{SimCLR Grad-CAM images}
\label{fig:c_cam}
\end{figure}

The effect of different augmentations introduced in SimCLR has been summarized in Table \ref{tab:simCLR}. The random gamma transform resulted in a slightly better performance (73.8\%) than the original SimCLR model (73\%). This can be attributed to the fact that gamma transform increased the contrast in images, thus making the fine-grained spots more prominent. The Grad-CAM image on Figure \ref{fig:c_gamma} showed that the model was able to localize the spots on the leaves without getting confused with the background. 

Using coarse dropout transform dropped the accuracy significantly to 48\%. Even though we ensured the fine-grained regions are not completely covered by the squares, the model was not able to localize the important spots from the unobstructed regions of the leaves (Figure \ref{fig:c_dropout}).

With random patch swapping, the model was localizing the spots on the leaf but at the same time, it was also using the unimportant background elements (Figure \ref{fig:c_swap}). Accuracy of 53\% was obtained.

We observed that the model was able to localize the spots despite the slight distortion introduced due to random patch swapping. Hence, we increased the randomness by introducing a Jigsaw shuffling. As observed from Figure \ref{fig:c_jigaw}, the model was not very confident about the spots on the leaves and was also highlighting some features from the background. It was able to achieve an accuracy of only 69.5\%.

With the intuition that random chopping and shuffling can make the fine-grained regions unidentifiable, we experimented with DCL-based ~\cite{8953746} Jigsaw shuffling. As can be observed from Figure \ref{fig:c_dcl}, the model was very confident and was able to localize the spots on the leaves, yet it was only able to reach an accuracy of 71.1\%.

Inspired by the idea that fine-grained datasets usually provide bounding boxes to localize the features, we experimented with using the original image and the fine-grained region as a positive pair. As observed from Figure \ref{fig:c_fine}, the model was unable to localize the discriminatory spots and was highlighting background elements while reaching an accuracy of only 54\%. Building upon this, we thought it would be interesting to see the effect of retaining the discriminatory fine-grained region and shuffling the remaining image region to induce randomness. This resulted in further deterioration of the downstream classification accuracy to 51\%. It might be the case that the model was using the leaf boundaries to locate the spots, while with the smartcropping most of the images were zoomed onto the spots and the leaf structure was lost. We believe SimCLR has a lot of potential due to its simple and intuitive nature and more can be explored.

\begin{minipage}{\linewidth}
\centering
\captionof{table}{Downstream validation accuracy using SimCLR. } \label{tab:simCLR}
\begin{threeparttable}
\begin{tabular}{ p{1.5in} p{0.85in} p{0.85in}}\toprule[1.5pt]
\bf Model & \bf Accuracy\\\midrule
Original & 0.730 \\
Coarse dropout & 0.480 \\
Gamma transform & 0.738 \\
Random patch swapping & 0.530 \\
Random Jigsaw (4x4) & 0.695 \\
DCL-based Jigsaw (3x3) & 0.711 \\
Fine-grained region crop & 0.548 \\
\bottomrule[1.25pt]
\end {tabular}\par
\begin{tablenotes}
    \item[*] We used $\tau$ = 0.5 and a batch size of 64 in the pretext task.
    \end{tablenotes}
\end{threeparttable}
\bigskip
\end{minipage}

\subsection{SRGAN}

On SRGAN generator, the best performing model used an input image size of 128 x 128 (Table \ref{tab:scaling_obs}). Our results showed roughly the same accuracy when no residual blocks were removed (64.5\%) and two residual blocks were removed (64.2\%). We chose the simpler model with two residual blocks removed to perform the subsequent experiments. Our results showed a dramatic increase in accuracy (up to 83.3\%) when dropouts were introduced in the classifier, with the best performing model using a dropout of 0.5.

Comparing the results of the generator and discriminator, we found that the discriminator initially performed better than the generator when changing the input image size and reducing the number of identical blocks from the baseline architecture. However, once dropouts were introduced, the generator significantly outperformed the discriminator.

\begin{minipage}{\linewidth}
\centering
\captionof{table}{Downstream validation accuracy using SRGAN} \label{tab:scaling_obs}
\begin{threeparttable}
\begin{tabular}{ p{1.5in} p{1.8in} p{0.85in} p{1.2in}}\toprule[1.5pt]
\bf Parameter & \bf Value & \bf Generator* & \bf Discriminator**\\
\bf & \bf & \bf Accuracy & \bf Accuracy\\\midrule
Image size & Original (crop 88 x 88) & 0.612 & 0.704\\
 & 256 x 256 (crop 88 x 88) & 0.621 & 0.713\\
 & 128 x 128 (crop 88 x 88) & 0.645 & 0.741\\
 & 88 x 88 (no cropping) & 0.619 & 0.645\\
\hline \\
 Depth & Remove the last $2^*$/$1^{**}$ blocks & 0.642 & 0.744\\
 of architecture & Remove the last $4^*$/$2^{**}$ blocks & 0.635 & 0.715\\
 & Remove the last $6^*$/$3^{**}$ blocks & 0.628 & 0.701\\
\hline \\
Dropout & 0.5 & 0.833 & 0.732\\
 & 0.7 & 0.800 & 0.746\\
 & 0.9 & 0.824 & 0.748\\
\bottomrule[1.25pt]
\end {tabular}\par
\begin{tablenotes}
    \item[*] The depth of ResNet was reduced in increments of two for the generator.
    \item[**] The depth of VGG was reduced in increments of one for the discriminator.
    \end{tablenotes}
\end{threeparttable}
\bigskip
\end{minipage}

\begin{figure}
\centering
\begin{subfigure}[b]{.2\linewidth}
\includegraphics[width=\linewidth]{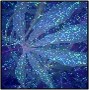}
\caption{Gen. (original)}\label{fig:gen_ori_gan}
\end{subfigure}
\begin{subfigure}[b]{.2\linewidth}
\includegraphics[width=\linewidth]{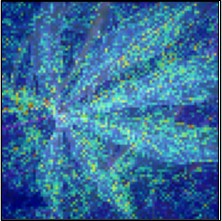}
\caption{128 x 128}\label{fig:gen_128}
\end{subfigure}
\begin{subfigure}[b]{.2\linewidth}
\includegraphics[width=\linewidth]{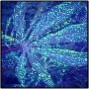}
\caption{Remove 2 blocks}\label{fig:gen_remove2resblocks_gan}
\end{subfigure}
\begin{subfigure}[b]{.2\linewidth}
\includegraphics[width=\linewidth]{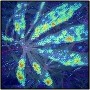}
\caption{Dropout 0.5}\label{fig:gen_dropout05_gan}
\end{subfigure}
\begin{subfigure}[b]{.2\linewidth}
\includegraphics[width=\linewidth]{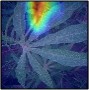}
\caption{Discr. (original)}\label{fig:disc_ori}
\end{subfigure}
\begin{subfigure}[b]{.2\linewidth}
\includegraphics[width=\linewidth]{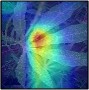}
\caption{128 x 128}\label{fig:disc_128}
\end{subfigure}
\begin{subfigure}[b]{.2\linewidth}
\includegraphics[width=\linewidth]{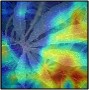}
\caption{Remove last block}\label{fig:disc_removelast1conv}
\end{subfigure}
\begin{subfigure}[b]{.2\linewidth}
\includegraphics[width=\linewidth]{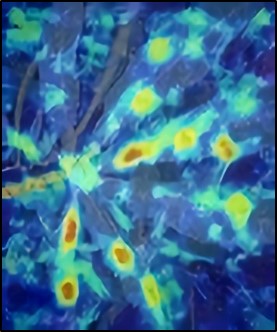}
\caption{Dropout 0.9}\label{fig:dcl}
\end{subfigure}
\caption{SRGAN Grad-CAM images for the best performing models for the different experiments using the generator (top) and discriminator (bottom). (a) and (e) are the results of the original generator and discriminator, respectively using the default parameters; (b) and (f) are the results for the best (i.e. highest accuracy) input size images; (c) and (g) are the results for the best baseline model depth; (d) and (h) are the results for best dropout hyperparameter.}
\label{fig:srgan_cam}
\end{figure}

We noted an interesting observation on the Grad-CAM images (Figure \ref{fig:srgan_cam}). Before dropouts were added, the discriminator was still able to learn and roughly base its classification decision on certain spots of the leaves, while the generator was not able to distinguish the leaves from the background at all. Once dropouts were added, both the generator and discriminator exhibited an improvement in performance, with the generator being able to supersede the discriminator and accurately determine the correct identifying features of the cassava leaves. This suggests that the models (particularly the generator) have great potential to be used as a feature extractor, but they need to be used in conjunction with a regularization technique such as dropout to separate the unimportant background features from the important ones.

\begin{figure}
\centering
\begin{subfigure}[b]{.2\linewidth}
\includegraphics[width=\linewidth]{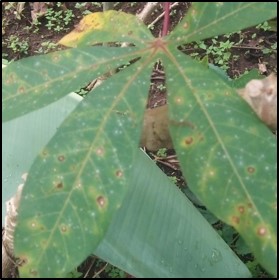}
\caption{}\label{fig:gen_ori_fail}
\end{subfigure}
\begin{subfigure}[b]{.2\linewidth}
\includegraphics[width=\linewidth]{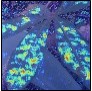}
\caption{}\label{fig:gen_128_fail}
\end{subfigure}
\begin{subfigure}[b]{.2\linewidth}
\includegraphics[width=\linewidth]{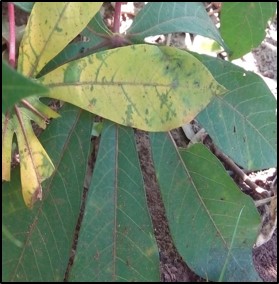}
\caption{}\label{fig:gen_remove2resblocks}
\end{subfigure}
\begin{subfigure}[b]{.2\linewidth}
\includegraphics[width=\linewidth]{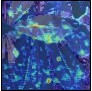}
\caption{}\label{fig:gen_dropout05}
\end{subfigure}
\caption{Success and failure cases of our best performing model. (a) and (c) are the original leaf images. (b) is an example of a success case where the model was clearly able to distinguish between the different leaf types. (d) is an example of a failure case where, in the presence of different-shaded cassava leaves, the model was probably not able to use color as a distinguishing factor.}
\label{fig:sfc}
\end{figure}

\section{Conclusion}

Fine-grained classification is an important problem owing to the number of applications in the real world. However, large annotated datasets are very expensive to obtain for training models. Hence, self-supervised learning was explored as a possible solution to exploit the freely available unlabelled fine-grained images. We experimented with Jigsaw as pretext task, SRGAN generator and SimCLR.

Comparing the best models for the different pretext tasks, we found that the Jigsaw task performed most poorly, followed by SimCLR, SRGAN discriminator, and SRGAN generator. We accomplished the highest downstream classification accuracy of 83\% (compared to the 88\% supervised fine-grained baseline model). We thus see a promising venue for self-supervised learning as a learning mechanism.

Our further investigation into the successes and failures of the models reveal that our models were likely able to learn the texture and shape of the leaves, but were probably not able to learn colors as a distinguishing factor between the classes (Figure \ref{fig:sfc}). For future research, we would be interested in exploring color as a pretext task. We would also consider repeating this experiment on other fine-grained datasets to verify the findings of our experiment.

\bibliographystyle{plain}
\bibliography{main.bib}

\centering \large \textbf{Appendix}

The following tables summarize the class-wise results obtained on each of the described models with different parameters.

\begin{table*}[h]
  \centering
  \begin{tabular}{ | l | l | l | l |}
    \toprule
\textbf{Class} & \textbf{Precision} & \textbf{Recall} & \textbf{F1-score}\\
 \hline
cbb &  0.71 & 0.75 & 0.73\\
cbsd & 0.82 & 0.91 & 0.87\\
cgm & 0.89 & 0.81 & 0.84\\
cmd & 0.83 & 0.95 & 0.94\\
healthy & 0.99 & 0.68 & 0.81\\
 \hline
  \end{tabular}
  \caption{Observed results for Baseline model}
  \label{tab:base}
\end{table*}

\begin{table*}[h]
  \centering
  \begin{tabular}{ | l | l | l | l |}
    \toprule
\textbf{Class} & \textbf{Precision} & \textbf{Recall} & \textbf{F1-score}\\
 \hline
cbb &  0.47 & 0.45 & 0.46\\
cbsd & 0.67 & 0.69 & 0.68\\
cgm & 0.23 & 0.55 & 0.32\\
cmd & 0.87 & 0.72 & 0.79\\
healthy & 0.7 & 0.64 & 0.67\\
 \hline
  \end{tabular}
  \caption{Observed results for Jigsaw as pretext task}
  \label{tab:jig}
\end{table*}

\begin{table*}[h]
  \centering
  \begin{tabular}{ | l | l | l | l |}
    \toprule
\textbf{Class} & \textbf{Precision} & \textbf{Recall} & \textbf{F1-score}\\
 \hline
 cbb & 0.52 & 0.51 & 0.52\\
 cbsd & 0.67 & 0.72 & 0.69\\
 cgm & 0.77 & 0.82 & 0.79\\
 cmd & 0.91 & 0.86 & 0.88\\
 healthy & 0.67 & 0.65 & 0.66\\
 \hline
  \end{tabular}
  \caption{Observed results for original SimCLR model downstream classification. Accuracy 73.1\%}
  \label{tab:original}
\end{table*}

\begin{table*}[h]
  \centering
  \begin{tabular}{ | l | l | l | l |}
    \toprule
\textbf{Class} & \textbf{Precision} & \textbf{Recall} & \textbf{F1-score}\\
 \hline
 cbb & 0.27 & 0.51 & 0.36\\
 cbsd & 0.52 & 0.47 & 0.49\\
 cgm & 0.32 & 0.47 & 0.38 \\
 cmd & 0.68 & 0.64 & 0.66\\
 healthy & 0.69& 0.50 & 0.58\\
 \hline
  \end{tabular}
  \caption{Observed results for SimCLR model with random patch swapping downstream classification. Accuracy 53\%}
  \label{tab:swap}
\end{table*}

\begin{table*}[h]
  \centering
  \begin{tabular}{ | l | l | l | l |}
    \toprule
\textbf{Class} & \textbf{Precision} & \textbf{Recall} & \textbf{F1-score}\\
 \hline
 cbb & 0.49 & 0.48 & 0.49\\
 cbsd & 0.56 & 0.73 & 0.64\\
 cgm & 0.75 & 0.74 & 0.74\\
 cmd & 0.86 & 0.87 & 0.86\\
 healthy & 0.72 & 0.61 & 0.66\\
 \hline
  \end{tabular}
  \caption{Observed results for SimCLR model with random Jigsaw 4x4 (Case 1 under \ref{sec:Jigsaw}) downstream classification. Accuracy 69.5\%}
  \label{tab:Jigsaw}
\end{table*}

\begin{table*}[h]
  \centering
  \begin{tabular}{ | l | l | l | l |}
    \toprule
\textbf{Class} & \textbf{Precision} & \textbf{Recall} & \textbf{F1-score}\\
 \hline
 cbb & 0.40 & 0.59 & 0.47\\
 cbsd & 0.74 & 0.59 & 0.65\\
 cgm & 0.75 & 0.76 & 0.76\\
 cmd & 0.86 & 0.86 & 0.86\\
 healthy & 0.66 & 0.65 & 0.65\\
 \hline
  \end{tabular}
  \caption{Observed results for SimCLR model with DCL 3x3 downstream classification. Accuracy 71.1\%}
  \label{tab:dcl}
\end{table*}

\begin{table*}[h]
  \centering
  \begin{tabular}{ | l | l | l | l |}
    \toprule
\textbf{Class} & \textbf{Precision} & \textbf{Recall} & \textbf{F1-score}\\
 \hline
 cbb & 0.25 & 0.43 & 0.31\\
 cbsd & 0.57 & 0.46 & 0.51\\
 cgm & 0.42 & 0.50 & 0.45\\
 cmd & 0.77 & 0.72 & 0.74\\
 healthy & 0.57 & 0.52 & 0.54\\
 \hline
  \end{tabular}
  \caption{Observed results for SimCLR model with fine grain region downstream classification. Accuracy 54.8\%}
  \label{tab:fine_region}
\end{table*}

\begin{table*}[h]
  \centering
  \begin{tabular}{ | l | l | l | l |}
    \toprule
\textbf{Class} & \textbf{Precision} & \textbf{Recall} & \textbf{F1-score}\\
 \hline
cbb & 0.54 & 0.64 & 0.59\\
cbsd & 0.75 & 0.62 & 0.68\\
cgm & 0.76 & 0.81 & 0.78\\
cmd & 0.91 & 0.87 & 0.89\\
healthy & 0.68 & 0.72 & 0.70\\
 \hline
  \end{tabular}
  \caption{Observed results for SimCLR model with gamma transform downstream classification. Accuracy 73.8\%}
  \label{tab:gamma}
\end{table*}

\begin{table*}[h]
  \centering
  \begin{tabular}{ | l | l | l | l |}
    \toprule
\textbf{Class} & \textbf{Precision} & \textbf{Recall} & \textbf{F1-score}\\
 \hline
cbb & 0.38 & 0.43 & 0.40\\
cbsd & 0.47 & 0.46 & 0.46\\
cgm & 0.35 & 0.40 & 0.37\\
cmd & 0.76 & 0.54 & 0.63\\
healthy & 0.35 & 0.49 & 0.41\\
 \hline
  \end{tabular}
  \caption{Observed results for SimCLR model with coarse dropout transform downstream classification. Accuracy 48\%}
  \label{tab:dropout48}
\end{table*}

\begin{table*}[h]
  \centering
  \begin{tabular}{ | l | l | l | l |}
    \toprule
\textbf{Class} & \textbf{Precision} & \textbf{Recall} & \textbf{F1-score}\\
 \hline
cbb & 0.44 & 0.13 & 0.20\\
cbsd & 0.60 & 0.67 & 0.64\\
cgm & 0.22 & 0.03 & 0.05\\
cmd & 0.63 & 0.92 & 0.75\\
healthy & 0.00 & 0.00 & 0.00\\
 \hline
  \end{tabular}
  \caption{Observed results for original SRGAN generator model without modification for downstream classification. Accuracy 61\%}
  \label{tab:dropout61}
\end{table*}

\begin{table*}[h]
  \centering
  \begin{tabular}{ | l | l | l | l |}
    \toprule
\textbf{Class} & \textbf{Precision} & \textbf{Recall} & \textbf{F1-score}\\
 \hline
cbb & 0.67 & 0.13 & 0.21\\
cbsd & 0.61 & 0.70 & 0.65\\
cgm & 0.29 & 0.12 & 0.17\\
cmd & 0.66 & 0.90 & 0.76\\
healthy & 0.00 & 0.00 & 0.00\\
 \hline
  \end{tabular}
  \caption{Observed results for SRGAN generator model with resize 256 x 256 for downstream classification. Accuracy 62\%}
  \label{tab:dropout62}
\end{table*}

\begin{table*}[h]
  \centering
  \begin{tabular}{ | l | l | l | l |}
    \toprule
\textbf{Class} & \textbf{Precision} & \textbf{Recall} & \textbf{F1-score}\\
 \hline
cbb & 0.46 & 0.19 & 0.27\\
cbsd & 0.64 & 0.72 & 0.68\\
cgm & 0.50 & 0.10 & 0.17\\
cmd & 0.66 & 0.93 & 0.77\\
healthy & 0.00 & 0.00 & 0.00\\
 \hline
  \end{tabular}
  \caption{Observed results for SRGAN generator model with resize 128 x 128 for downstream classification. Accuracy 64\%}
  \label{tab:dropout64}
\end{table*}

\begin{table*}[h]
  \centering
  \begin{tabular}{ | l | l | l | l |}
    \toprule
\textbf{Class} & \textbf{Precision} & \textbf{Recall} & \textbf{F1-score}\\
 \hline
cbb & 0.50 & 0.13 & 0.20\\
cbsd & 0.58 & 0.66 & 0.62\\
cgm & 0.33 & 0.06 & 0.10\\
cmd & 0.65 & 0.93 & 0.76\\
healthy & 0.33 & 0.04 & 0.08\\
 \hline
  \end{tabular}
  \caption{Observed results for SRGAN generator model with resize 88 x 88 for downstream classification. Accuracy 62\%}
  \label{tab:dropout621}
\end{table*}

\begin{table*}[h]
  \centering
  \begin{tabular}{ | l | l | l | l |}
    \toprule
\textbf{Class} & \textbf{Precision} & \textbf{Recall} & \textbf{F1-score}\\
 \hline
cbb & 0.16 & 0.45 & 0.23\\
cbsd & 0.72 & 0.64 & 0.68\\
cgm & 0.10 & 0.50 & 0.17\\
cmd & 0.93 & 0.66 & 0.77\\
healthy & 0.00 & 0.00 & 0.00\\
 \hline
  \end{tabular}
  \caption{Observed results for SRGAN generator model after removing the last two residual blocks (for downstream classification). Accuracy 64\%}
  \label{tab:dropout641}
\end{table*}

\begin{table*}[h]
  \centering
  \begin{tabular}{ | l | l | l | l |}
    \toprule
\textbf{Class} & \textbf{Precision} & \textbf{Recall} & \textbf{F1-score}\\
 \hline
cbb & 0.50 & 0.09 & 0.16\\
cbsd & 0.65 & 0.70 & 0.67\\
cgm & 0.56 & 0.07 & 0.13\\
cmd & 0.64 & 0.95 & 0.76\\
healthy & 0.00 & 0.00 & 0.00\\
 \hline
  \end{tabular}
  \caption{Observed results for SRGAN generator model after removing the last four residual blocks (for downstream classification). Accuracy 64\%}
  \label{tab:dropout_gan}
\end{table*}

\begin{table*}[h]
  \centering
  \begin{tabular}{ | l | l | l | l |}
    \toprule
\textbf{Class} & \textbf{Precision} & \textbf{Recall} & \textbf{F1-score}\\
 \hline
cbb & 0.56 & 0.28 & 0.38\\
cbsd & 0.77 & 0.87 & 0.82\\
cgm & 0.81 & 0.76 & 0.78\\
cmd & 0.88 & 0.95 & 0.92\\
healthy & 0.73 & 0.48 & 0.58\\
 \hline
  \end{tabular}
  \caption{Observed results for SRGAN generator model with dropout probability of 0.5 (downstream classification). Accuracy 83\%}
  \label{tab:dropout_83}
\end{table*}

\begin{table*}[h]
  \centering
  \begin{tabular}{ | l | l | l | l |}
    \toprule
\textbf{Class} & \textbf{Precision} & \textbf{Recall} & \textbf{F1-score}\\
 \hline
cbb & 0.50 & 0.53 & 0.52\\
cbsd & 0.78 & 0.79 & 0.79\\
cgm & 0.84 & 0.64 & 0.73\\
cmd & 0.84 & 0.95 & 0.89\\
healthy & 0.30 & 0.48 & 0.41\\
 \hline
  \end{tabular}
  \caption{Observed results for SRGAN generator model with dropout probability of 0.7 (downstream classification). Accuracy 80\%}
  \label{tab:dropout_80}
\end{table*}

\begin{table*}[h]
  \centering
  \begin{tabular}{ | l | l | l | l |}
    \toprule
\textbf{Class} & \textbf{Precision} & \textbf{Recall} & \textbf{F1-score}\\
 \hline
cbb & 0.50 & 0.28 & 0.36\\
cbsd & 0.77 & 0.84 & 0.81\\
cgm & 0.88 & 0.73 & 0.80\\
cmd & 0.86 & 0.96 & 0.91\\
healthy & 0.80 & 0.52 & 0.63\\
 \hline
  \end{tabular}
  \caption{Observed results for SRGAN generator model with dropout probability of 0.9 (downstream classification). Accuracy 82\%}
  \label{tab:dropout_82}
\end{table*}

\end{document}